\documentclass[11pt]{article}
\usepackage{acl2014}
\usepackage{times}
\usepackage{url}
\usepackage{latexsym}
\usepackage{amsmath}
\usepackage{dsfont}
\usepackage{multirow}
\usepackage{balance}

\title{Credibility Adjusted Term Frequency: A Supervised Term Weighting Scheme for Sentiment Analysis and Text Classification}

\author{Yoon Kim \\ New York University \\ {\tt yhk255@nyu.edu} \And Owen Zhang \\  {\tt zhonghua.zhang2006@gmail.com}}

\date{}

\begin{document}
\maketitle
\begin{abstract}
We provide a simple but novel supervised weighting scheme for adjusting term frequency in \emph{tf-idf} for sentiment analysis and text classification. We compare our method to baseline weighting schemes and find that it outperforms them on multiple benchmarks. The method is robust and works well on both snippets and longer documents.
\end{abstract}

\section{Introduction}
Baseline discriminative methods for text classification usually involve training a linear classifier over bag-of-words (BoW) representations of documents. In BoW representations (also known as Vector Space Models), a document is represented as a vector where each entry is a count (or binary count) of tokens that occurred in the document. Given that some tokens are more informative than others, a common technique is to apply a weighting scheme to give more weight to discriminative tokens and less weight to non-discriminative ones. Term frequency-inverse document frequency (\emph{tf-idf}) \cite{Salton:1983} is an unsupervised weighting technique that is commonly employed. In \emph{tf-idf}, each token $i$ in document $d$ is assigned the following weight,
\begin{equation}
w_{i,d} = tf_{i,d} \cdot \log \frac{N}{df_i}
\end{equation}
where $tf_{i,d}$ is the number of times token $i$ occurred in document $d$, $N$ is the number of documents in the corpus, and $df_i$ is the number of documents in which token $i$ occurred.

Many supervised and unsupervised variants of \emph{tf-idf} exist (Debole and Sebastiani \shortcite{Debole:2003}; Martineau and Finin \shortcite{Martineau:2009}; Wang and Zhang \shortcite{Wang:2013}). The purpose of this paper is not to perform an exhaustive comparison of existing weighting schemes, and hence we do not list them here. Interested readers are directed to Paltoglou and Thelwall \shortcite{Paltoglou:2010} and Deng et al. \shortcite{Deng:2014} for comprehensive reviews of the different schemes. 

In the present work, we propose a simple but novel supervised method to adjust the term frequency portion in \emph{tf-idf} by assigning a credibility adjusted score to each token. We find that it outperforms the traditional unsupervised \emph{tf-idf} weighting scheme on multiple benchmarks. The benchmarks include both snippets and longer documents. We also compare our method against Wang and Manning \shortcite{Sida:2012}'s Naive-Bayes Support Vector Machine (NBSVM), which has achieved state-of-the-art results (or close to it) on many datasets, and find that it performs competitively against NBSVM. We additionally find that the traditional \emph{tf-idf} performs competitively against other, more sophisticated methods when used with the right scaling and normalization parameters.
\section{The Method}
Consider a binary classification task. Let $C_{i,k}$ be the count of token $i$ in class $k$, with $k \in \{-1,1\}$. Denote $C_{i}$ to be the count of token $i$ over both classes, and $y^{(d)}$ to be the class of document $d$. For each occurrence of token $i$ in the training set, we calculate the following,
\begin{equation}
s_{i}^{(j)} = \begin{cases} \frac{C_{i,1}}{C_i} &\mbox{, if } y^{(d)} = 1 \\
\frac{C_{i,-1}}{C_i} &\mbox{, if } y^{(d)} = -1 \end{cases}
\end{equation}
Here, $j$ is the $j$-th occurrence of token $i$. Since there are $C_i$ such occurrences, $j$ indexes from 1 to $C_i$. We assign a score to token $i$ by,
\begin{equation}
\hat{s_i} = \frac{1}{C_i} \sum_{j=1}^{C_i} s_{i}^{(j)}
\end{equation}
Intuitively, $\hat{s}_i$ is the average likelihood of making the correct classification given token $i$'s occurrence in the document, if $i$ was the only token in the document. In a binary classification case, this reduces to,
\begin{equation}
\hat{s_i} = \frac{C_{i,1}^2 + C_{i,-1}^2}{C_i^2}
\end{equation}
Note that by construction, the support of $\hat{s_i}$ is $[0.5,1]$.
\subsection{Credibility Adjustment}
Suppose $\hat{s_i} = \hat{s_j} = 0.75$ for two different tokens $i$ and $j$, but $C_i = 5$ and $C_j = 100$. Intuition suggests that $\hat{s_j}$ is a more credible score than $\hat{s_i}$, and that $\hat{s_i}$ should be shrunk towards the population mean. Let $\hat{s}$ be the (weighted) population mean. That is,
\begin{equation}
\hat{s} = \sum_i\frac{C_i \cdot \hat{s}_i}{C}
\end{equation}
where $C$ is the count of all tokens in the corpus. We define \emph{credibility adjusted score} for token $i$ to be,
\begin{equation}
\overline{s}_i = \frac{C_{i,1}^2 + C_{i,-1}^2 + \hat{s}\cdot\gamma}{C_i^2 + \gamma}
\end{equation}
where $\gamma$ is an additive smoothing parameter. If $C_{i,k}$'s are small, then $\overline{s}_i \approx \hat{s}$ (otherwise, $\overline{s}_i \approx \hat{s}_i$). This is a form of Buhlmann credibility adjustment from the actuarial literature \cite{Buhlmann:2005}. 
We subsequently define $\overline{tf}$, the \emph{credibility adjusted term frequency}, to be,
\begin{equation}
\overline{tf}_{i,d} = (0.5 + \overline{s}_{i}) \cdot tf_{i,d}
\end{equation}
and $tf$ is replaced with $\overline{tf}$. That is, 
\begin{equation}
w_{i,d} = \overline{tf}_{i,d} \cdot \log \frac{N}{df_i}
\end{equation}
We refer to above as \emph{cred-tf-idf} hereafter.
\subsection{Sublinear Scaling}
It is common practice to apply sublinear scaling to $tf$. A word occurring (say) ten times more in a document is unlikely to be ten times as important. Paltoglou and Thelwall \shortcite{Paltoglou:2010} confirm that sublinear scaling of term frequency results in significant improvements in various text classification tasks. We employ logarithmic scaling, where $tf$ is replaced with $\log(tf) + 1$. For our method, $\overline{tf}$ is simply replaced with $\log(\overline{tf}) + 1$. We found virtually no difference in performance between log scaling and other sublinear scaling methods (such as augmented scaling, where $tf$ is replaced with $0.5 + \frac{0.5 + tf}{\max tf}$).

\subsection{Normalization}
Using normalized features resulted in substantial improvements in performance versus using un-normalized features. We thus use $\hat{\mathbf{x}}^{(d)} = \mathbf{x}^{(d)} / ||\mathbf{x}^{(d)}||_2$ in the SVM, where $\mathbf{x}^{(d)}$ is the feature vector obtained from \emph{cred-tf-idf} weights for document $d$.
\subsection{Naive-Bayes SVM (NBSVM)}
Wang and Manning \shortcite{Sida:2012} achieve excellent (sometimes state-of-the-art) results on many benchmarks using binary Naive Bayes (NB) log-count ratios as features in an SVM. In their framework, 
\begin{equation}
w_{i,d} = \mathbf{1}\{tf_{i,d}\} \log \frac{(df_{i,1} + \alpha)/\sum_i(df_{i,1} + \alpha)}{(df_{i,-1} + \alpha)/\sum_i(df_{i,-1} + \alpha)}
\end{equation}
where $df_{i,k}$ is the number of documents that contain token $i$ in class $k$, $\alpha$ is a smoothing parameter, and $\mathbf{1}\{\cdot\}$ is the indicator function equal to one if $tf_{i,d} > 0$ and zero otherwise.  As an additional benchmark, we implement NBSVM with $\alpha=1.0$ and compare against our results.\footnote{Wang and Manning \shortcite{Sida:2012} use the same $\alpha$ but they differ from our NBSVM in two ways. One, they use $l_2$ hinge loss (as opposed to $l_1$ loss in this paper). Two, they interpolate NBSVM weights with Multivariable Naive Bayes (MNB) weights to get the final weight vector. Further, their tokenization is slightly different. Hence our NBSVM results are not directly comparable. We list their results in table 2.}
\section{Datasets and Experimental Setup}
We test our method on both long and short text classification tasks, all of which were used to establish baselines in Wang and Manning \shortcite{Sida:2012}. Table 1 has summary statistics of the datasets.
\begin{table}
\center
\begin{tabular}{|c|c|c|c|c|}
\hline
\textbf{Dataset} & \textbf{Length} & \textbf{Pos} & \textbf{Neg} & \textbf{Test} \\ \hline
PL-sh & 21 & 5331 & 5331 & CV \\ 
PL-sub & 24 & 5000 & 5000 & CV \\ \hline
PL-2k & 746 & 1000 & 1000 & CV \\ 
IMDB & 231 & 12.5k & 12.5k & 25k \\
AthR & 355 & 480 & 377 & 570 \\ 
XGraph & 276 & 584 & 593 & 784 \\ \hline
\end{tabular}
\caption{Summary statistics for the datasets. Length is the average number of unigram tokens (including punctuation) per document. Pos/Neg is the number of positive/negative documents in the training set. Test is the number of documents in the test set (CV means that there is no separate test set for this dataset and thus a 10-fold cross-validation was used to calculate errors).}
\end{table}
The snippet datasets are:
\begin{itemize}
  \item \textbf{PL-sh}: Short movie reviews with one sentence per review. Classification involves detecting whether a review is positive or negative. \cite{Pang:2005}.\footnote{https://www.cs.cornell.edu/people/pabo/movie-review-data/. All the PL datasets are available here.}
  \item \textbf{PL-sub}: Dataset with short subjective movie reviews and objective plot summaries. Classification task is to detect whether the sentence is objective or subjective. \cite{Pang:2004}. 
\end{itemize}
And the longer document datasets are:
\begin{itemize}
  \item \textbf{PL-2k}: 2000 full-length movie reviews that has become the de facto benchmark for sentiment analysis \cite{Pang:2004}.
  \item \textbf{IMDB}: 50k full-length movie reviews (25k training, 25k test), from IMDB \cite{Maas:2011}.\footnote{http://ai.stanford.edu/~amaas/data/sentiment/index.html}
  \item \textbf{AthR, XGraph}: The 20-Newsgroup dataset, 2nd version with headers removed.\footnote{http://people.csail.mit.edu/jrennie/20Newsgroups} Classification task is to classify which topic a document belongs to. AthR: alt.atheism vs religion.misc, XGraph: comp.windows.x vs comp.graphics.
\end{itemize}

\subsection{Support Vector Machine (SVM)}
For each document, we construct the feature vector $\mathbf{x}^{(d)}$ using weights obtained from \emph{cred-tf-idf} with log scaling and $l_2$ normalization.  For \emph{cred-tf-idf}, $\gamma$ is set to 1.0. NBSVM and \emph{tf-idf} (also with log scaling and $l_2$ normalization) are used to establish baselines. Prediction for a test document is given by
\begin{equation}
y^{(d)} = \mbox{sign } (\mathbf{w}^T\mathbf{x}^{(d)} + b)
\end{equation}
In all experiments, we use a Support Vector Machine (SVM) with a linear kernel and penalty parameter of $C = 1.0$. For the SVM, $\mathbf{w}$, $b$ are obtained by minimizing,
\begin{equation}
\mathbf{w}^T\mathbf{w} + C\sum_{d=1}^N \max (0, 1 - y^{(d)}(\mathbf{w}^T\mathbf{x}^{(d)} + b))
\end{equation}
using the LIBLINEAR library \cite{Fan:2008}.
\subsection{Tokenization}
We lower-case all words but do not perform any stemming or lemmatization. We restrict the vocabulary to all tokens that occurred at least twice in the training set.

\begin{table*}[ht]
\center
\begin{tabular}{|l|l|c c|c c c c|}
\hline
& \textbf{Method} & \textbf{PL-sh} & \textbf{PL-sub} & \textbf{PL-2k} & \textbf{IMDB} & \textbf{AthR} & \textbf{XGraph} \\ \hline
& tf-idf-uni & 77.1 & 91.5 & 88.1 & 88.6 & 85.8 & 88.4 \\
& tf-idf-bi & 78.0 & 92.3 & 89.2 & 90.9 & 86.5 & 88.0 \\
Our & cred-tfidf-uni & 77.5 & 91.8 & 88.7 & 88.8 & 86.5 & 89.8 \\
results & cred-tfidf-bi & 78.6 & \textbf{92.8} & \textbf{89.7} & \textbf{91.3} & \textbf{87.4} & 88.9 \\
& NBSVM-uni & 75.5 & 89.9 & 87.0 & 85.9 & 86.7 & 88.5 \\
& NBSVM-bi & 76.0 & 90.5 & 89.5 & 90.5 & 86.7 & 88.1 \\ \hline
& MNB-uni & 77.9 & 92.6 & 83.5 & 83.6 & 85.0 & 90.0 \\
Wang \& & MNB-bi & \textbf{79.0} & \textbf{\underline{93.6}} & 85.9 & 86.6 & 85.1 & \textbf{91.2} \\
Manning & NBSVM-uni & 78.1 & 92.4 & 87.8 & 88.3 &\textbf{\underline{87.9}} & \textbf{\underline{91.2}} \\
& NBSVM-bi & \textbf{\underline{79.4}} & \textbf{93.2} & \textbf{89.5} & \textbf{91.2} & \textbf{87.7} & \textbf{90.7} \\ \hline
& Appr. Tax.\text{*} & - & - & 90.2 & - & - & - \\
& Str. SVM\text{*} & - & - & 92.4 & - & - & - \\
& aug-tf-mi & - & - & 87.8 & 88.0& - & - \\
Other & Disc. Conn. & - & - & - & \textbf{\underline{91.4}} & - & - \\
results & Word Vec.\text{*} & - & 88.6 & 88.9 & 88.9 & - & - \\ 
& LLR & - & - & \textbf{\underline{90.4}} & - & - & - \\ 
& RAE & 77.7 & - & - & - & - & - \\ 
& MV-RNN & \textbf{79.0} & - & - & - & - & - \\ \hline
\end{tabular}
\caption{Results of our method (\emph{cred-tf-idf}) against baselines (\emph{tf-idf}, NBSVM), using unigrams and bigrams. \emph{cred-tf-idf} and \emph{tf-idf} both use log scaling and $l_2$ normalization. Best results (that do not use external sources) are underlined, while top three are in bold. Rows 7-11 are MNB and NBSVM results from Wang and Manning \shortcite{Sida:2012}. Our NBSVM results are not directly comparable to theirs (see footnote 1). Methods with \text{*} use external data or software. \textbf{Appr. Tax}: Uses appraisal taxonomies from \emph{WordNet} \cite{Whitelaw:2005}. \textbf{Str. SVM}: Uses \emph{OpinionFinder} to find objective versus subjective parts of the review \cite{Yes:2010}. \textbf{aug-tf-mi}: Uses augmented term-frequency with mutual information gain \cite{Deng:2014}. \textbf{Disc. Conn.}: Uses discourse connectors to generate additional features \cite{Trivedi:2013}. \textbf{Word Vec.}: Learns sentiment-specific word vectors to use as features combined with BoW features \cite{Maas:2011}. \textbf{LLR}: Uses log-likelihood ratio on features to select features \cite{Aue:2005}. \textbf{RAE}: Recursive autoencoders \cite{Socher:2011}. \textbf{MV-RNN}: Matrix-Vector Recursive Neural Networks \cite{Socher:2012}.}
\end{table*}
\section{Results and Discussion}
For PL datasets, there are no separate test sets and hence we use 10-fold cross validation (as do other published results) to estimate errors. The standard train-test splits are used on IMDB and Newsgroup datasets.
\subsection{\emph{cred-tf-idf} outperforms \emph{tf-idf}}
Table 2 has the comparison of results for the different datasets. Our method outperforms the traditional \emph{tf-idf} on all benchmarks for both unigrams and bigrams. While some of the differences in performance are significant at the 0.05 level (e.g. IMDB), some are not (e.g. PL-2k). The Wilcoxon signed ranks test is a non-parametric test that is often used in cases where two classifiers are compared over multiple datasets \cite{Demsar:2006}. The Wilcoxon signed ranks test indicates that the overall outperformance is significant at the $<$0.01 level.
\subsection{NBSVM outperforms \emph{cred-tf-idf}}
\emph{cred-tf-idf} did not outperform Wang and Manning \shortcite{Sida:2012}'s NBSVM (Wilcoxon signed ranks test $p$-value = 0.1). But it did outperform our own implementation of NBSVM, implying that the extra modifications by Wang and Manning \shortcite{Sida:2012} (i.e. using squared hinge loss in the SVM and interpolating between NBSVM and MNB weights) are important contributions of their methodology. This was especially true in the case of shorter documents, where our uninterpolated NBSVM performed significantly worse than their interpolated NBSVM.
\subsection{\emph{tf-idf} still performs well}
We find that \emph{tf-idf} still performs remarkably well with the right scaling and normalization parameters. Indeed, the traditional \emph{tf-idf} outperformed many of the more sophisticated methods that employ distributed representations (Maas et al. \shortcite{Maas:2011}; Socher et al. \shortcite{Socher:2011}) or other weighting schemes (Martineau and Finin \shortcite{Martineau:2009}; Deng et al. \shortcite{Deng:2014}).
\section{Conclusions and Future Work}
In this paper we presented a novel supervised weighting scheme, which we call \emph{credibility adjusted term frequency}, to perform sentiment analysis and text classification. Our method outperforms the traditional \emph{tf-idf} weighting scheme on multiple benchmarks, which include both snippets and longer documents. We also showed that \emph{tf-idf} is competitive against other state-of-the-art methods with the right scaling and normalization parameters.

From a performance standpoint, it would be interesting to see if our method is able to achieve even better results on the above tasks with proper tuning of the $\gamma$ parameter. Relatedly, our method could potentially be combined with other supervised variants of \emph{tf-idf}, either directly or through ensembling, to improve performance further.
\balance

\end{document}